\documentclass{article}

\usepackage[nonatbib, final]{nips_2017}

\usepackage[utf8]{inputenc} 
\usepackage[T1]{fontenc}    
\usepackage{hyperref}       
\usepackage{url}            
\usepackage{booktabs}       
\usepackage{amsfonts}       
\usepackage{nicefrac}       
\usepackage{microtype}      

\usepackage{epsfig}
\usepackage{graphicx}
\usepackage{amsmath}
\usepackage{amssymb}

\usepackage{color}
\usepackage{array, floatrow, tabularx, makecell, booktabs}
\setcellgapes{3pt}

\usepackage{wrapfig}
\usepackage{verbatim}

\title{WRPN: Wide Reduced-Precision Networks}

\author{
Asit Mishra \hspace{1.25mm} Eriko Nurvitadhi \hspace{1.25mm} Jeffrey J Cook \hspace{1.25mm} Debbie Marr \\
Accelerator Architecture Lab, Intel
}

\begin{document}

\maketitle

\begin{abstract}
For computer vision applications, prior works have shown the
efficacy of reducing numeric precision
of model parameters (network weights) in deep neural networks.
Activation maps, however, occupy a large memory footprint
during both the training and inference step when using mini-batches
of inputs. One way to reduce this large memory footprint is to reduce the
precision of activations. However, past works have shown that
reducing the precision of activations hurts model accuracy.
We study schemes to {\it train} networks from scratch using reduced-precision
activations without hurting accuracy. We reduce the precision
of activation maps (along with model parameters) and
increase the number of filter maps in a layer, and find that this
scheme matches or surpasses the accuracy of the baseline
full-precision network.
As a result, one can significantly improve the execution efficiency (e.g. reduce dynamic memory
footprint, memory bandwidth and computational energy) and speed up the training
and inference process with appropriate hardware support.
We call our scheme WRPN - wide reduced-precision
networks. We report results and show
that WRPN scheme is better than previously reported accuracies on ILSVRC-12
dataset while being computationally less expensive compared to previously
reported reduced-precision networks.

\end{abstract}

\section{Introduction}

A promising approach to lower the compute and memory requirements of
convolutional deep-learning
workloads is through the use of low numeric precision algorithms.
Operating in lower precision mode reduces computation as
well as data movement and storage requirements.
Due to such efficiency benefits, there are many existing works which propose
low-precision deep neural networks (DNNs)~\cite{INQ, NN-FewMult,
LogNN, StochasticRounding, GoogleLowPrecision}, even down to 2-bit
ternary mode~\cite{TTQ, TWN, AALTWN} and 1-bit binary mode~\cite{DoReFa, BNN,
XNORNET, BWN, FINN}. However, the majority of existing works in low-precision
DNNs sacrifice accuracy over the baseline full-precision networks.
Further, most prior works target reducing the precision of the model
parameters (network weights). This primarily benefits the inference
step only when batch sizes are small.


To improve both execution efficiency and accuracy of
low-precision networks, we reduce
both the precision of activation maps and
model parameters and increase the number of filter maps in a layer.
We call networks using this scheme wide reduced-precision networks
(WRPN) and find that this scheme compensates or surpasses the
accuracy of the baseline full-precision network.
Although the number of raw compute operations increases as we increase the
number of filter maps in a layer,
the compute bits required per operation is now a fraction
of what is required when using full-precision operations
(e.g. going from FP32 AlexNet to 4-bits precision
and doubling the number of filters increases the
number of compute operations by 4x, but each operation is 8x more efficient than FP32).

WRPN offers better accuracies,
while being computationally less expensive compared to
previously reported reduced-precision networks.
We report results
on AlexNet~\cite{AlexNet}, batch-normalized Inception~\cite{InceptionBN}, and
ResNet-34~\cite{ResNet} on ILSVRC-12~\cite{AlexNet} dataset.
We find 4-bits to be sufficient for training deep and wide models
while achieving similar or better accuracy than baseline network.
With 4-bit activation and
2-bit weights, we find the accuracy to be at-par with baseline full-precision.
Making the networks wider and operating with
1-bit precision, we close the
accuracy gap between previously report
binary networks and show state-of-the art
results for ResNet-34 (69.85\% top-1 with 2x wide) and
AlexNet (48.04\% top-1 with 1.3x wide). To the best of our knowledge,
our reported accuracies with binary networks (and even 4-bit precision)
are highest to date.

Our reduced-precision quantization scheme is hardware friendly
allowing for efficient hardware implementations.
To this end, we evaluate efficiency benefits of low-precision
operations (4-bits to 1-bits) on Titan X GPU, Arria-10 FPGA and ASIC. We see that FPGA and ASIC can deliver significant
efficiency gain over FP32 operations (6.5x to 100x), while GPU cannot take advantage of very low-precision operations.


\section{Motivation for reduced-precision activation maps}

\begin{figure}[!htb]
\begin{center}
   \includegraphics[width=0.90\linewidth]{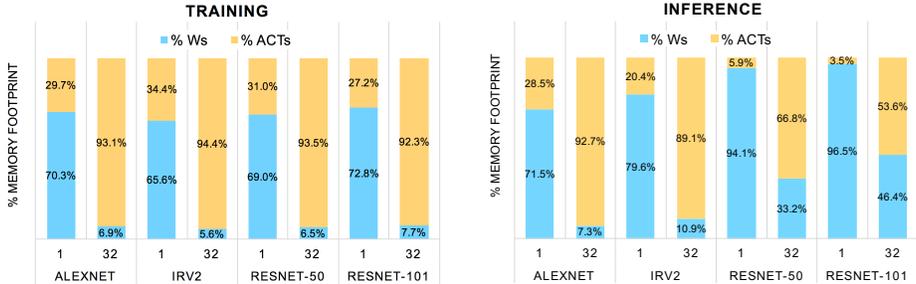}
\end{center}
   \caption{\small Memory footprint of activations (ACTs) and
   weights (W) during training and inference for
   mini-batch sizes 1 and 32.}
\label{fig:MemoryFootprint}
\end{figure}

While most prior works proposing reduced-precision networks work
with low precision
weights (e.g.~\cite{BNN, TTQ, DoReFa, AALTWN, TWN, BWN, FINN}),
we find that activation maps occupy a larger
memory footprint when using mini-batches of inputs.
Using mini-batches of inputs is typical in
training of DNNs and cloud-based batched inference~\cite{TPU}.
Figure~\ref{fig:MemoryFootprint} shows memory footprint of
activation maps and filter maps as batch size changes for
4 different networks (AlexNet, Inception-Resnet-v2~\cite{IRv2},
ResNet-50 and ResNet-101) during the training and inference steps.

\begin{figure}[!htb]
\begin{center}
   \includegraphics[width=0.70\textwidth]{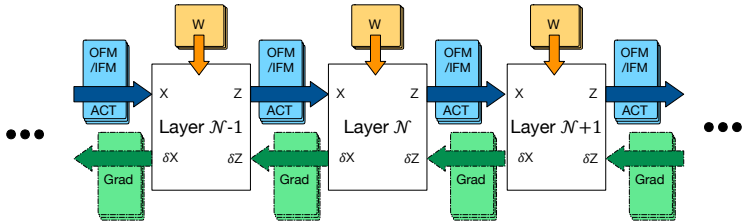}
\end{center}
   \caption{\small Memory requirements of a feed forward
   convolutional deep neural network. Orange boxes denote weights (W), blue boxes
   are activations (ACT) and green boxes are gradient-maps (Grad).}
\label{fig:DNNLayers}
\end{figure}

As batch-size increases, because of filter reuse across batches of
inputs, activation maps occupy significantly
larger fraction of memory compared to the filter weights.
This aspect is illustrated in Figure~\ref{fig:DNNLayers}
which shows the memory requirements of a canonical feed-forward DNN for a
hardware accelerator based system (e.g. GPU, FPGA, PCIe connected ASIC device, etc.).
During training, the sum of all the activation maps
(ACT) and weight tensors (W) are allocated in device memory for forward
pass along with memory for gradient maps during backward propagation.
The total memory requirements for training phase is the sum of memory
required for the activation maps, weights and the maximum of
input gradient maps ($\delta$Z)
and maximum of back-propagated gradients ($\delta$X).
During inference, memory is allocated for input (IFM)
and output feature maps (OFM)
required by a single layer, and
these memory allocations are reused for other layers.
The total memory allocation during inference is then the
maximum of IFM and maximum of OFM required across all
the layers plus the sum of all W-tensors.
At batch sizes 128 and more, activations
start to occupy more than 98\% of total memory footprint during training.

Overall, reducing precision of activations and weights reduces memory footprint,
bandwidth and storage while also simplifying the requirements
for hardware to efficiently support these operations.


\section{WRPN scheme and studies on AlexNet}

Based on the observation that activations occupy more memory
footprint compared to weights, we reduce the precision of activations
to speed up training and inference steps as well as cut down on
memory requirements. However, a straightforward reduction
in precision of activation maps leads to significant reduction in
model accuracy~\cite{DoReFa, XNORNET}.

We conduct a sensitivity study where we reduce precision of
activation maps and model weights for AlexNet running ILSVRC-12 dataset
and train the network from scratch.
Table~\ref{table:AlexNet1x-accuracy} reports our findings.
Top-1 single-precision (32-bits weights and activations) accuracy is 57.2\%.
The accuracy with binary weights and activations is 44.2\%.
This is similar to what is reported in~\cite{XNORNET}.
$32b A$ and $2b W$ data-point in this table is
using TTQ technique~\cite{TTQ}.
All other data points are collected using our quantization scheme
(described later in Section~\ref{sec:quantizationScheme}),
all the runs have same hyper-parameters and training is carried
out for the same number of epochs as baseline network.
To be consistent with results reported in prior works,
we do not quantize weights and activations of the first
and last layer.

We find that, in general, reducing the precision of activation
maps and weights hurts model accuracy.
Further, reducing precision of activations hurts model accuracy much
more than reducing precision of the filter parameters.
We find TTQ
to be quite effective on AlexNet in that one
can lower the precision of weights to 2b (while activations are still FP32)
and not lose accuracy.
However, we did not find this scheme to be effective for other
networks like ResNet or Inception.

\begin{table}[!htb]
\scriptsize
  \floatsetup{floatrowsep=qquad, captionskip=2pt}
  \begin{floatrow}[2]
    \ttabbox%
    {\begin{tabularx}{0.475\textwidth}{XXXXXX}
      \toprule
            & 32b A & 8b A & 4b A & 2b A & 1b A   \\
      \cmidrule(lr){2-2}\cmidrule(lr){3-3}\cmidrule(lr){4-4}\cmidrule(lr){5-5}\cmidrule(lr){6-6}
      32b W	& 57.2 & 54.3 & 54.4 & 52.7 & --      \\
      8b W	& --    & 54.5 & 53.2 & 51.5 & --     \\
      4b W	& --    & 54.2 & 54.4 & 52.4 & --     \\
      2b W	& 57.5 & 50.2 & 50.5 & 51.3 & --      \\
      1b W	& 56.8 & --    & --    & --	  & 44.2  \\
      \bottomrule
      \end{tabularx}}
    {\caption[AlexNet top-1 accuracy]{AlexNet top-1 validation set
    accuracy \% as precision of activations (A) and weight(W) changes.
    All results are with end-to-end training of the network from scratch.
    $-$ is a data-point we did not experiment for.}
    \label{table:AlexNet1x-accuracy}}
  \hfill%
  \ttabbox%
  {\begin{tabularx}{0.475\textwidth}{XXXXXX}
    \toprule
          & 32b A & 8b A & 4b A & 2b A & 1b A \\
    \cmidrule(lr){2-2}\cmidrule(lr){3-3}\cmidrule(lr){4-4}
    \cmidrule(lr){5-5}\cmidrule(lr){6-6}
    32b W	& 60.5  & 58.9 & 58.6 & 57.5 & 52.0   \\
    8b W	& --    & 59.0 & 58.8 & 57.1 & 50.8   \\
    4b W	& --    & 58.8 & 58.6 & 57.3 & --     \\
    2b W	& --    & 57.6 & \textcolor{blue}{57.2} & 55.8 & --     \\
    1b W	& --    & --   & --   &  --	 & 48.3   \\
    \bottomrule
    \end{tabularx}}
  {\caption[AlexNet 2x-wide top-1 validation set accuracy]
  {AlexNet 2x-wide
  top-1 validation set accuracy \% as precision of activations
  (A) and weights (W) changes.}
  \label{table:AlexNet2x-accuracy}}
  \end{floatrow}
\end{table}%


To re-gain the model accuracy while working with reduced-precision
operands, we increase the number of filter maps in a layer.
Although the number of raw compute operations increase
with widening the filter maps in a layer,
the bits required per compute operation is now a fraction of
what is required when using full-precision operations.
As a result, with appropriate hardware support, one can significantly
reduce the dynamic memory requirements, memory bandwidth,
computational energy and speed up the training and inference process.

Our widening of filter maps is inspired from
Wide ResNet~\cite{WideResNets} work where the depth
of the network is reduced and
width of each layer is increased (the operand precision is still FP32).
Wide ResNet requires a re-design of the
network architecture. In our work, we maintain the depth
parameter same as baseline network but widen the filter maps.
We call our approach WRPN - wide reduced-precision networks.
In practice, we find this scheme to be very simple and effective
- starting with a baseline network architecture, one can change the
width of each filter map without changing any other
network design parameter or hyper-parameters.
Carefully reducing precision and simultaneously
widening filters keeps the total compute cost of the network
under or at-par with baseline cost.\footnote{Compute cost is the product of the number of FMA operations
and the sum of width of the activation and weight operands.}

Table~\ref{table:AlexNet2x-accuracy} reports the
accuracy of AlexNet when we double the number of filter
maps in a layer. With doubling of filter maps, AlexNet with
4-bits weights and 2-bits activations exhibits accuracy at-par with
full-precision networks. Operating with 4-bits weights and
4-bits activations surpasses the baseline accuracy by 1.44\%.
With binary weights and activations we better the
accuracy of XNOR-NET~\cite{XNORNET} by 4\%.

When doubling the number of filter maps, AlexNet's raw compute operations
grow by 3.9x compared to the baseline full-precision network,
however by using reduced-precision operands the overall
compute complexity is a fraction of the baseline.
For example, with 4b operands for weights and activations and
2x the number of filters, reduced-precision AlexNet is just 49\%
of the total compute cost of the full-precision baseline
(compute cost comparison is shown
in Table~\ref{table:AlexNet2x-Cost}).

\begin{table}[!htb]
  \floatsetup{floatrowsep=qquad, captionskip=2pt}
  \begin{floatrow}
    \ttabbox%
    {\begin{tabularx}{0.65\textwidth}{XXXXXX}
      \toprule
            & 32b A & 8b A & 4b A & 2b A & 1b A \\
      \cmidrule(lr){2-2}\cmidrule(lr){3-3}\cmidrule(lr){4-4}
      \cmidrule(lr){5-5}\cmidrule(lr){6-6}
      32b W	& 3.9x  & 2.4x & 2.2x & 2.1x & 2.0x   \\
      8b W	& 2.4x  & 1.0x & 0.7x & 0.6x & 0.6x   \\
      4b W	& 2.2x  & 0.7x & 0.5x & 0.4x & 0.3x   \\
      2b W	& 2.1x  & 0.6x & 0.4x & 0.2x & 0.2x   \\
      1b W	& 2.0x  & 0.6x & 0.3x & 0.2x & 0.1x   \\
      \bottomrule
      \end{tabularx}}
    {\caption[AlexNet 2x-wide compute cost]{Compute cost of
    AlexNet 2x-wide vs. 1x-wide as precision of activations (A)
    and weights (W) changes.}
    \label{table:AlexNet2x-Cost}}
  \end{floatrow}
\end{table}%

We also experiment with other widening factors.
With 1.3x widening of filters
and with 4-bits of activation precision one can go as
low as 8-bits of weight precision while still being at-par with
baseline accuracy.
With 1.1x wide filters, at least 8-bits weight and
16-bits activation precision
is required for accuracy to match baseline
full-precision 1x wide accuracy.
Further, as Table~\ref{table:AlexNet2x-Cost} shows,
when widening filters
by 2x, one needs to lower precision to at least 8-bits so that
the total compute cost is not more than baseline compute cost.
Thus, there is a trade-off between widening and reducing the
precision of network parameters.

In our work, we trade-off higher number of raw
compute operations with aggressively
reducing the precision of the operands involved in these operations
(activation maps and filter weights) while not
sacrificing the model accuracy.
Apart from other benefits of reduced precision activations
as mentioned earlier,
widening filter maps also improves the efficiency of underlying
GEMM calls for convolution operations since compute
accelerators are typically more efficient on a single kernel
consisting of parallel computation on large data-structures
as opposed to many small sized kernels~\cite{WideResNets}.

\section{Studies on deeper networks}

We study how our scheme applies to deeper networks. For this, we study
ResNet-34~\cite{ResNet} and batch-normalized Inception~\cite{InceptionBN} and find similar
trends, particularly that 2-bits weight and 4-bits activations continue to provide
at-par accuracy as baseline.
We use TensorFlow~\cite{tensorflow2015-whitepaper}
and tensorpack~\cite{tensorpack} for all our
evaluations and use ILSVRC-12 train and val dataset for
analysis.\footnote{We will open-source our implementation of
reduced-precision AlexNet, ResNet and batch-normalized Inception networks.}

\subsection{ResNet}

ResNet-34 has 3x3 filters in each of its modular
layers with shortcut connections being 1x1.
The filter bank width changes from 64 to 512 as depth
increases. We use the pre-activation variant of ResNet
and the baseline top-1 accuracy of our ResNet-34 implementation
using single-precision 32-bits data format is 73.59\%.
Binarizing weights and activations for all layers except the
first and the last layer in this network gives top-1 accuracy of 60.5\%.
For binarizing ResNet we did not re-order any layer (as is
done in XNOR-NET). We used the same hyper-parameters and
learning rate schedule as the baseline network.
As a reference, for ResNet-18, the gap between XNOR-NET (1b weights
and activations) and full-precision
network is 18\% \cite{XNORNET}.
It is also interesting to note that top-1 accuracy of
single-precision AlexNet (57.20\%) is lower than the top-1
accuracy of binarized ResNet-34 (60.5\%).

\begin{table}[!htb]
  \floatsetup{floatrowsep=qquad, captionskip=2pt}
  \begin{floatrow}
    \ttabbox%
    {\begin{tabularx}{0.70\textwidth}{XXXX}
      \toprule
        Width & Precision & Top-1 Acc. \%    & Compute cost \\
      \cmidrule(lr){1-4}
      1x wide & 32b A, 32b W  & 73.59 & 1x    \\
              & 1b A, 1b W    & 60.54 & 0.03x \\
      \cmidrule(lr){1-4}
      2x wide & 4b A, 8b W    & 74.48 & 0.74x \\
              & 4b A, 4b W    & 74.52 & 0.50x \\
              & \textcolor{blue}{4b A, 2b W}    & \textcolor{blue}{73.58} & 0.39x \\
              & 2b A, 4b W    & 73.50 & 0.39x \\
              & 2b A, 2b W    & 73.32 & 0.27x \\
              & \textcolor{blue}{1b A, 1b W}    & \textcolor{blue}{69.85} & 0.15x \\
      \cmidrule(lr){1-4}
      3x wide & 1b A, 1b W   & 72.38 & 0.30x  \\
      \bottomrule
      \end{tabularx}}
    {\caption[ResNet]{ResNet-34 top-1 validation accuracy \%
    and compute cost as precision of activations (A)
    and weights (W) varies.}
    \label{table:ResnetData}}
  \end{floatrow}
\end{table}%

We experimented with doubling number of filters in
each layer
and reduce the precision of activations and weights.
Table~\ref{table:ResnetData} shows the results of our analysis.
Doubling the number of filters and
4-bits precision for both weights and activations beats the baseline
accuracy by 0.9\%. 4-bits activations and 2-bits (ternary)
weights has top-1 accuracy at-par with baseline. Reducing precision to
2-bits for both weights and activations degrades accuracy
by only 0.2\% compared to baseline.

Binarizing the weights and activations with 2x wide filters
has a top-1 accuracy of 69.85\%. This is just 3.7\% worse
than baseline full-precision network while being only
15\% of the cost of the baseline network.
Widening the filters by 3x and binarizing the
weights and activations reduces this gap to 1.2\% while
the 3x wide network is 30\% the cost of the
full-precision baseline network.

Although 4-bits precision seems to be enough for wide networks,
we advocate for 4-bits activation precision and
2-bits weight precision.
This is because with ternary weights one can get rid of the
multipliers and use adders instead.
Additionally, with this
configuration there is no loss of accuracy. Further,
if some accuracy
degradation is tolerable, one can even go to binary circuits for
efficient hardware implementation while saving 32x in bandwidth
for each of weights and activations compared to
full-precision networks. All these gains can be realized with
simpler hardware implementation and lower compute cost
compared to baseline networks.


To the best of our knowledge, our ResNet binary
and ternary (with 2-bits or 4-bits activation) top-1 accuracies
are state-of-the-art results in the literature
including unpublished technical reports (with similar
data augmentation~\cite{PCLpaper1}).

\subsection{Batch-normalized Inception}

We applied WRPN scheme to batch-normalized Inception
network~\cite{InceptionBN}.
This network includes batch normalization of all layers
and is a variant of GoogleNet~\cite{GoogleNet} where the 5x5 convolutional
filters are replaced by two 3x3 convolutions with up to 128 wide filters.
Table~\ref{table:InceptionBNData} shows the results of our analysis.
Using 4-bits activations and 2-bits weight and doubling the number of
filter banks in the network produces a model that is almost at-par
in accuracy with the baseline single-precision network (0.02\% loss in
accuracy). Wide network with binary weights and
activations is within 6.6\% of the full-precision
baseline network.

\begin{table}[!htb]
  \floatsetup{floatrowsep=qquad, captionskip=2pt}
  \begin{floatrow}
    \ttabbox%
    {\begin{tabularx}{0.70\textwidth}{XXXX}
      \toprule
        Width & Precision & Top-1 Acc. \%    & Compute cost \\
      \cmidrule(lr){1-4}
      1x wide & 32b A, 32b W  & 71.64 & 1x    \\
      \cmidrule(lr){1-4}
      2x wide & 4b A, 4b W    & 71.63 & 0.50x \\
              & \textcolor{blue}{4b A, 2b W}    & \textcolor{blue}{71.61} & 0.38x \\
              & 2b A, 2b W    & 70.75 & 0.25x \\
              & 1b A, 1b W    & 65.02 & 0.13x \\
      \bottomrule
      \end{tabularx}}
    {\caption[BNInception]{Batch-normalized Inception
    top-1 validation accuracy \% and compute cost
    as precision of activations (A) and weights (W) varies.}
    \label{table:InceptionBNData}}
  \end{floatrow}
\end{table}%


\section{Hardware friendly quantization scheme}
\label{sec:quantizationScheme}

We adopt the straight-through estimator (STE) approach
in our work~\cite{STE1}.
When quantizing a real number to $k$-bits, the
ordinality of the set of quantized numbers is
$2^k$. Mathematically, this small and finite set
would have zero gradients with respect to its inputs.
STE method circumvents this problem by defining an operator
that has arbitrary forward and backward operations.


Prior works using the STE approach define operators
that quantize the weights based on the expectation of the weight
tensors. For instance, TWN~\cite{TWN} uses a threshold and a scaling
factor for each layer to quantize weights to ternary domain.
In TTQ~\cite{TTQ}, the scaling factors are learned
parameters. XNOR-NET binarizes the weight tensor by computing
the sign of the tensor values and then scaling by the mean of the
absolute value of {\it each} output channel of weights.
DoReFa uses a single scaling factor across the entire layer.
For quantizing weights to $k$-bits, where $k > 1$, DoReFa uses:

\begin{equation} \label{eq2}
w^{k} = 2 * quantize_{k}(\frac{tanh(w_{i})}{2 * max(\mid tanh(w_i) \mid)} + \frac{1}{2}) - 1)
\end{equation}

Here $w_{k}$ is the k-bit quantized version of inputs
$w_{i}$ and quantize$_{k}$ is a quantization function that
quantizes a floating-point number $w_{i}$ in the range $[0,1]$
to a $k$-bit number in the same range. The transcendental
$tanh$ operation
constrains the weight value to lie in between $-1$ and $+1$.
The affine transformation post quantization brings the range to $[-1,1]$.

We build on these approaches and propose a much simpler scheme.
For quantizing weight tensors we first hard
constrain the values to lie within the range $[-1,1]$
using min-max operation (e.g. tf.clip$\_$by$\_$val when
using Tensorflow~\cite{tensorflow2015-whitepaper}).
For quantizing activation tensor values, we constrain the
values to lie within the range $[0,1]$. This step is followed
by a quantization step where a real number is
quantized into a $k$-bit number. This is given as, for $k > 1$:

\begin{equation} \label{eq3}
w_{k} = \frac{1}{2^{k-1} - 1}round((2^{k-1} - 1)* w_{i})
\hspace{3mm} \quad\text{and}\quad \hspace{3mm}
a_{k} = \frac{1}{2^{k} - 1}round((2^{k} - 1)* a_{i})
\end{equation}


Here $w_{i}$ and $a_{i}$ are input real-valued weights
and activation tensor
and $w_{k}$ and $a_{k}$ are their quantized versions.
One bit is reserved for sign-bit in case of weight values,
hence the use of $2^{k-1}$ for these quantized values.
Thus, weights can be stored and interpreted using signed data-types
and activations using un-signed data-types.
With appropriate affine transformations, the convolution
operations (the bulk of the compute operations in the network
during forward pass) can be done using quantized values
(integer operations in hardware) followed by scaling with
floating-point constants (this scaling operation can be
done in parallel with
the convolution operation in hardware).
When $k = 1$, for binary weights
we use the BWN approach~\cite{BWN} where the binarized
weight value is computed based on the sign of input value followed by
scaling with the mean of absolute values. For binarized activations
we use the formulation in Eq.~\ref{eq3}.
We do not quantize the gradients and maintain the weights
in reduced precision format.

For convolution operation when using WRPN, the forward pass during
training (and the inference step) involves matrix multiplication
of $k$-bits signed and $k$-bits unsigned operands.
Since gradient values are in 32-bits floating-point
format, the backward pass involves a matrix multiplication
operation using 32-bits and $k$-bits operand
for gradient and weight update.

When $k > 1$, the hard clipping of tensors to a range maps
efficiently to min-max comparator units in hardware as opposed to
using transcendental operations which are long latency operations.
TTQ and DoRefa schemes involve division operation and
computing a maximum value in the input tensor. Floating-point
division operation is expensive in hardware and computing the
maximum in a tensor is an $O(n)$ operation.
Additionally, our quantization parameters are static and do
not require any learning or involve back-propagation
like TTQ approach. We avoid each of these costly
operations and propose a simpler
quantization scheme (clipping followed by rounding).


\subsection{Efficiency improvements of reduced-precision operations on GPU, FPGA and ASIC}

In practice, the effective performance and
energy efficiency one could achieve on a low-precision
compute operation highly depends on the hardware that runs these
operations. We study the
efficiency of low-precision operations on various
hardware targets – GPU, FPGA, and ASIC.

\begin{figure}[!htb]
\begin{center}
   \includegraphics[width=0.9\textwidth]{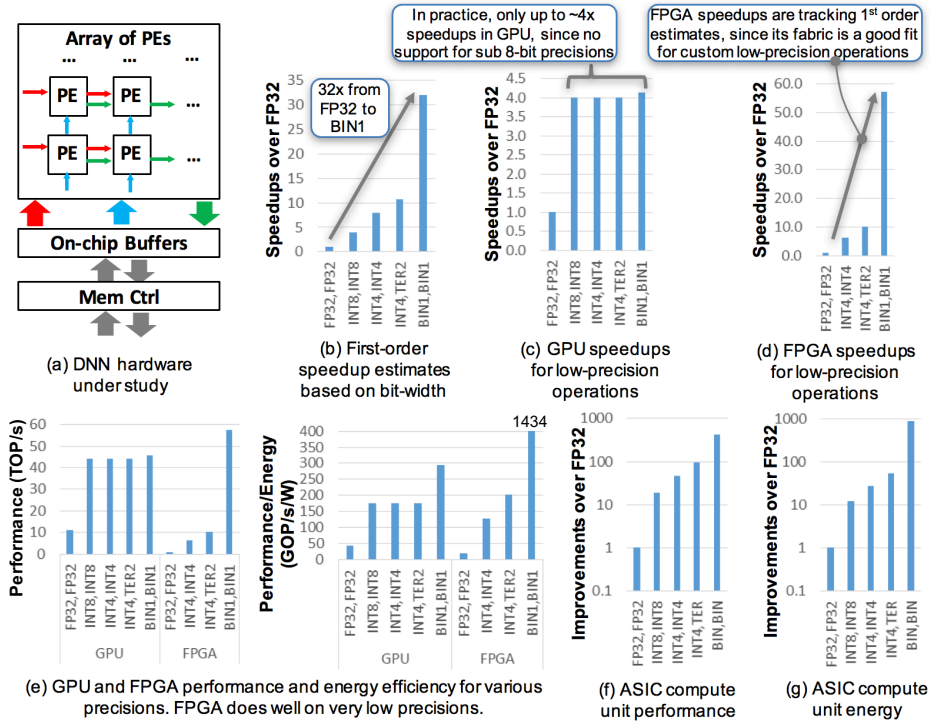}
\end{center}
\caption{\small Efficiency improvements from low-precision
operations on GPU, FPGA and ASIC.}
\label{HardwareAnalysis}
\end{figure}

For GPU, we evaluate WRPN on Nvidia Titan X Pascal
and for FPGA we use Intel Arria-10. We collect
performance numbers from both previously
reported analysis~\cite{FPGAvsGPU}
as well as our own experiments.
For FPGA, we implement a DNN accelerator architecture
shown in Figure~\ref{HardwareAnalysis}(a). This is a
prototypical accelerator design used in various
works (e.g., on FPGA~\cite{FPGAvsGPU} and ASIC such as TPU~\cite{TPU}).
The core of the accelerator consists of a systolic
array of processing elements (PEs) to perform matrix and
vector operations, along with on-chip buffers, as well as
off-chip memory management unit. The PEs can be configured to
support different precision -- (FP32, FP32),
(INT4, INT4), (INT4, TER2), and (BIN1, BIN1).
The (INT4, TER2) PE operates on ternary (+1,0,-1) values
and is optimized to include only an adder since there is no need
for a multiplier in this case. The binary (BIN1, BIN1) PE is
implemented using XNOR and bitcount. 
Our RTL design targets Arria-10 1150 FPGA.
For our ASIC study, we synthesize the PE design using Intel 14 nm
process technology to obtain area and energy estimates.



Figure~\ref{HardwareAnalysis}(b) - (g) summarize our analysis.
Figure~\ref{HardwareAnalysis}(b) shows the efficiency improvements
using first-order estimates where the efficiency is computed based on
number of
bits used in the operation.
With this method
we would expect (INT4, INT4) and (BIN1, BIN1) to be
8x and 32x more efficient, respectively, than (FP32, FP32).
However, in practice 
the efficiency gains from reducing precision depend on
whether the underlying hardware can take advantage of such low-precisions.

Figure~\ref{HardwareAnalysis}(c) shows performance
improvement on Titan X GPU
for various low-precision operations relative to
FP32. In this case, GPU can only achieve up to $\sim$4x
improvements in performance over FP32 baseline.
This is because GPU only provides first-class support for
INT8 operations, and is not able to take advantage of the
lower INT4, TER2, and BIN1 precisions. On the contrary,
FPGA can take advantage of such low precisions, since they are
amenable for implementations on the FPGA’s reconfigurable fabric.

Figure~\ref{HardwareAnalysis}(d) shows that the performance improvements
from (INT4, INT4), (INT4, TER2), and (BIN1, BIN1)
track well with the first-order
estimates from Figure~\ref{HardwareAnalysis}(b).
In fact, for (BIN1, BIN1), FPGA improvements exceed
the first-order estimate.
Reducing
the precision simplifies the design of compute units
and lower buffering requirements
on FPGA board. Compute-precision
reduction leads to significant improvement in
throughput due to smaller hardware designs
(allowing more parallelism) and shorter
circuit delay (allowing higher frequency).
Figure~\ref{HardwareAnalysis}(e)
shows the performance and performance/Watt of
the reduced-precision operations on GPU and FPGA.
FPGA performs quite well on very low precision operations.
In terms of performance/watt, FPGA does better than
GPU on (INT4, INT4) and lower precisions.

ASIC allows for a truly customized hardware implementation.
Our ASIC study provides insights to the upper bound of
the efficiency benefits possible from low-precision operations.
Figure~\ref{HardwareAnalysis}(f) and ~\ref{HardwareAnalysis}(g)
show improvement in performance and energy efficiency of the
various low-precision ASIC PEs relative to baseline
FP32 PE. As the figures show, going to lower precision offers
2 to 3 orders of magnitude efficiency improvements.

In summary, FPGA and ASIC are well suited for our WRPN approach.
At 2x wide, our WRPN approach requires ~4x more total
operations than the original network. However, for INT4 or lower precision,
each operation is ~6.5x or better in efficiency than FP32 for
FPGA and ASIC. Hence, WRPN delivers an overall efficiency win.

\section{Related work}

Reduced-precision DNNs is an active research area.
Reducing precision of weights for efficient
inference pipeline has been very well studied.
Works like Binary connect (BC)~\cite{BWN}, Ternary-weight
networks (TWN)~\cite{TWN}, fine-grained ternary
quantization~\cite{PCLpaper1} and INQ~\cite{INQ} target
reducing the precision of network weights
while still using full-precision activations. Accuracy is
almost always degraded when quantizing the weights.
For AlexNet on Imagenet, TWN loses 5\% top-1 accuracy.
Schemes like INQ,~\cite{ResilientDNNs} and \cite{PCLpaper1}
do fine-tuning to quantize the network weights and do not
sacrifice accuracy as much but are not applicable for
training networks from scratch. INQ shows promising results
with 5-bits of precision.

XNOR-NET~\cite{XNORNET}, BNN~\cite{BNN}, DoReFa~\cite{DoReFa}
and TTQ~\cite{TTQ} target training as well. While TTQ
targets weight quantization only, most works targeting
activation quantization 
hurt accuracy. XNOR-NET approach reduces top-1 accuracy by 12\%
and DoReFa by 8\% when quantizing both weights and activations
to 1-bit (for AlexNet on ImageNet). Further, XNOR-NET requires
re-ordering of layers for its scheme to work.
Recent work in~\cite{GrahamLowPrec} targets low-precision
activations and reports accuracy within 1\% of baseline
with 5-bits precision and logarithmic (with base $\sqrt{2}$)
quantization. With fine-tuning this
gap can be narrowed to be within 0.6\% but not all
layers are quantized.

Non-multiples of two for operand values
introduces hardware inefficiency in that
memory accesses are no longer DRAM or cache-boundary aligned
and end-to-end run-time performance
aspect is unclear when using complicated
quantization schemes.
We target end-to-end training
and inference, using very simple quantization
method and aim for reducing precision without
any loss in accuracy. To the best of our knowledge,
our work is the first to
study reduced-precision deep and wide networks,
and show accuracy at-par with baseline for as low a
precision as 4-bits activations and 2-bits weights.
We report state of the art accuracy for wide
binarized AlexNet and ResNet while still being lower
in compute cost.

\section{Conclusions}

We present the Wide Reduced-Precision Networks (WRPN) scheme for DNNs.
In this scheme, the numeric precision of both weights
and activations are significantly reduced without loss of
network accuracy.  This result is in contrast to many
previous works that find reduced-precision activations
to detrimentally impact accuracy; specifically, we find
that 2-bit weights and 4-bit activations are
sufficient to match baseline accuracy across many networks
including AlexNet, ResNet-34 and batch-normalized Inception.
We achieve this result with a new quantization
scheme and by increasing the number of filter maps in
each reduced-precision layer to compensate for
the loss of information capacity induced by reducing the precision.

We motivate this work with our observation that
full-precision activations contribute significantly more to
the memory footprint than full-precision weight parameters
when using mini-batch sizes common during
training and cloud-based inference; furthermore, by reducing the precision of
both activations and weights the compute complexity is
greatly reduced (40\% of baseline for 2-bit weights and 4-bit activations).

The WRPN quantization scheme and computation on low
precision activations and weights is hardware friendly
making it viable for
deeply-embedded system deployments as well as in
cloud-based training and inference servers with
compute fabrics for low-precision.
We compare Titan X GPU, Arria-10 FPGA and ASIC
implementations using WRPN and show our scheme
increases performance and energy-efficiency for
iso-accuracy across each.  Overall, reducing the
precision allows custom-designed compute units and
lower buffering requirements to provide significant
improvement in throughput.

\pagebreak
{\small
\bibliographystyle{ieee}
\bibliography{nips_refs}
}

\end{document}